\documentclass{article}
\usepackage{ijcai23}

\usepackage{times}
\usepackage{soul}
\usepackage{url}
\usepackage[hidelinks]{hyperref}
\usepackage[utf8]{inputenc}
\usepackage[small]{caption}
\usepackage{graphicx}
\usepackage{amsmath}
\usepackage{amssymb}
\usepackage{amsthm}
\usepackage{booktabs}
\usepackage{algorithm}
\usepackage{algorithmic}
\usepackage[switch]{lineno}
\usepackage{bbding}
\usepackage{multirow}
\usepackage{bm}

\usepackage{caption}
\usepackage{graphicx}
\usepackage{float} 
\usepackage{subcaption}
\usepackage{makecell}

\usepackage{xcolor}

\newcommand{\eg}{\textit{e.g.}}
\newcommand{\ie}{\textit{i.e.}}

\title{Structure-Aware Group Discrimination with Adaptive-View Graph Encoder: A Fast Graph Contrastive Learning Framework}

 \author{
 Zhenshuo Zhang$^1$
 \and
 Yun Zhu$^1$\and
 Haizhou Shi$^2$\And
Siliang Tang$^1$
 \affiliations
 $^1$Zhejiang University\\
 $^2$Rutgers University
 \emails
 \{zs.zhang, zhuyun\_dcd\}@zju.edu.cn,
 haizhou.shi@rutgers.edu,
 siliang@zju.edu.cn
 }

\begin{document}

\maketitle
\begin{abstract}
 
Albeit having gained significant progress lately, large-scale graph representation learning remains expensive to train and deploy for two main reasons: (i) the repetitive computation of multi-hop message passing and non-linearity in graph neural networks (GNNs); (ii) the computational cost of complex pairwise contrastive learning loss. Two main contributions are made in this paper targeting this twofold challenge: we first propose an adaptive-view graph neural encoder (AVGE) with a limited number of message passing to accelerate the forward pass computation, and then we propose a structure-aware group discrimination (SAGD) loss in our framework which avoids inefficient pairwise loss computing in most common GCL and improves the performance of the simple group discrimination. By the framework proposed, we manage to bring down the training and inference cost on various large-scale datasets by a significant margin (250x faster inference time) without loss of the downstream-task performance.
  
\end{abstract}
\section{introduction}

Graph Neural Networks~(GNNs) have shown superiority in dealing with graph-structured data, such as social networks{fan2019graph}, traffic networks~\cite{derrow2021eta}, and molecular graphs~\cite{xie2021mars}. In real-world scenarios, however, large-scale graph data often lack human-annotated labels, which creates a huge barrier for the traditional supervised learning paradigm. To conquer this limitation, self-supervised graph representation learning methods have been widely studied, among which Graph Contrastive Learning~(GCL) is dominant due to its ability to learn robust and generalizable representations for the downstream tasks~\cite{DGI,MVGRL,BGRL}. In GCL, the graph data encoder is trained to produce the representation space that minimizes the distance between the semantically invariant perturbed instances, e.g., sub-graphs created with mild augmentation, and maximizes the distance between irrelevant instances, e.g., randomly sampled sub-graphs. 

Although proven to be effective, the existing GCL methods have limitations in real-world large-scale graph data applications: since they typically require large amounts of time and computational resources to deploy. For one thing, the most common GNN encoders utilize multi-hop information in graphs by multi-layer message passing in every calculation step, which leads to large computational costs for both training and inference. And for another, the predominant pairwise constrictive loss is not efficient enough and takes lots of time until convergence. In the supervised setting, there are several works addressing the first problem by reducing the number of parameterized message passing~\cite{SGC} or distilling the trained GNN to Multi-Layer Perceptron~(MLP) to improve inference speed~\cite{nosmog}. As for the second problem, various techniques have been studied such as simplifying positive and negative sample construction process~\cite{SUGRL}, and removing the negative sample generation process~\cite{BGRL,BYOL,shi2020run}. However, those works didn't explore the application of a more efficient encoder in GCL and discuss the relationship between encoder and pretext tasks.

\begin{figure}
  \includegraphics[width=\linewidth]{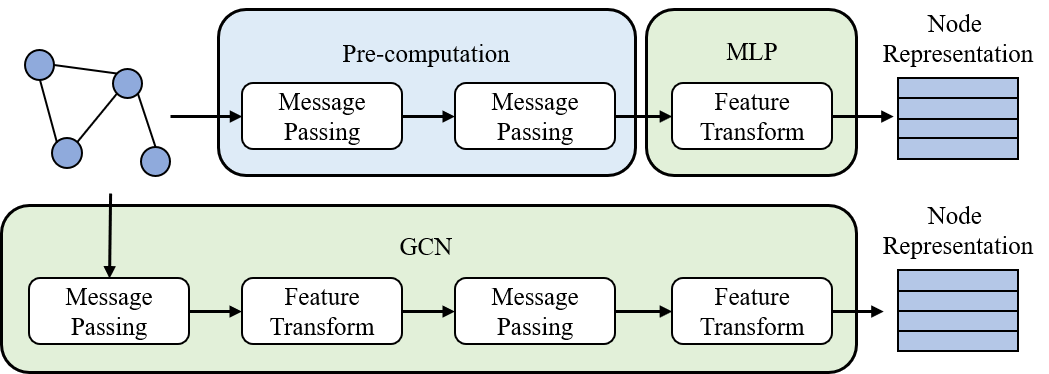}
  \caption{The architecture of separating 2-hop message passing and feature transform (above), compare to 2-layer GCN architecture (below). GCN will degrade to MLP if feature message passing is pred-computed and removed in each GCN layer.}
  \label{fig:precomputation}
\end{figure}

In this paper, we propose a novel GCL framework (AVGE-SAGD) to tackle the aforementioned two challenges. It contains an adaptive-view graph encoder~(AVGE) that achieves higher training and inference speed than the GNN counterparts, and a structure-aware group discrimination~(SAGD) module that increases the speed of the pretext contrastive task training.
In the AVGE, instead of using GNN, we first perform a limited number of message-passing to generate a multi-view feature vector that consists of multi-hop features. During training, the multi-view features are adaptively input to the encoder. And then we use an MLP encoder to further learn high-level representations for the pretext task. This encoder is significantly more efficient since it separates message passing and feature encoding and strictly controls the number of both operations.
In the SAGD module, we first introduce the group discrimination loss to avoid inefficient pairwise contrastive loss computation~\cite{rethinking}. Considering that the AVGE views the input multi-hop vector as a collection of independent features, it will lose the structural information of the original graph. Therefore to empower AVGE in the scenario of self-supervised representation learning, a novel structure prediction loss is added. It requires the encoded feature to further divide the graph into meaningful groups. By this extra topological constraint, we manage to prevent performance degradation and even achieve performance improvement.

To summarize, our contributions are as follows: 
\begin{itemize}
    \item We propose a graph encoder that adaptively utilizes multi-hop neighbor information, and separates the message passing from the encoder calculation procedure to save the repeated message passing calculation steps in the traditional GNN encoders, thereby improving the training speed and inference speed of our framework. 

    \item We propose a novel structure-aware group discrimination~(SAGD) module for GCL. It is built on graph group discrimination and further requires the encoder to subdivide the group into topology-based mini-groups so that the pre-trained model preserves more structural information and achieves better generalization ability for downstream tasks.
    \item Experiments on various node classification datasets showcase the effectiveness of our framework in terms of training and inference efficiency and downstream-task performance. Especially on large-scale graph data, our method achieves comparable performance with less training time and 250x faster inference time.
\end{itemize}

\section{Related Work}

Our framework involves two aspects: GNN encoder architecture and GCL methods. In this chapter, we introduce several previous works, discuss their limitations in GCL and propose our ideas for improvement.

\subsection{GNN Architecture}

Architecture design is an important part of GNN research. The most mainstream GNN structures are designed based on message passing, the most widely known of which is GCN~\cite{GCN}. There are several works that tried to accelerate the computing speed of GNN by separating the message passing phase and feature calculation as shown in Figure \ref{fig:precomputation}. The most known architecture is SGC~\cite{SGC}, which removes the non-linearity calculation between GCN layers and simplified it to linear transform, showing that parameters-free linear message passing can achieve similar performance to GCNs.
NAFS~\cite{NAFSzhang2022nafs} present learning-free node-adaptive feature smoothing, assign fixed weights for features in different hops by computing the distance from the aggregated features to the extreme over-smoothed features, and combine the features in different hops by summation.

Some studies have shown that under a certain design, the joint use of different hop features can enhance expressive ability. ASGC~\cite{ASGCchanpuriya2022simplified} uses the linear regression method to fit the raw features by constructing a linear combination of different hop features, thereby solving the problem of heterophily graph node classification. GCN-PND~\cite{heimpact} updates graph topology based on the similarity between the local neighborhood distribution of nodes and designing extensible aggregation from multi-hop neighbors.

These methods of separating message passing and feature computation are all applied in supervised scenarios, and a combination method such as summation is used for multi-hop information to keep data scale. We explore the separation of message passing and feature computation GNN architectures in self-supervised scenarios.

\subsection{Self-Supervised Graph Representation Learning}

Self-supervised learning was first proposed in the computer vision area and has quickly received widespread attention in the community of graph learning due to its excellent performance in scenarios with few labeled training data. There are three levels of contrastive learning in the GCL field: graph-graph level~\cite{you2020graph}, node-graph level~\cite{DGI,MVGRL}, and node-node level~\cite{GMI,GRACE,zhu2022rosa}. Among those methods, we focus on the node-graph level since our framework is also a kind of node-graph level contrastive learning. DGI~\cite{DGI} obtains the graph-level representation by applying a readout function on the graph and maximizes the mutual information between the patch and the graph representation to perform node-graph level graph contrastive learning. MVGRL~\cite{MVGRL} uses multi-view constructiveness to extend the idea of DGI and borrow the idea from graph diffusion networks~\cite{klicpera2019diffusion} to improve the performance. The training loss of DGI can be simplified into a binary classification loss which is empirically and theoretically proven in~\cite{rethinking}. The training scheme in~\cite{rethinking} is coined as \emph{Group Discrimination} which can implement efficient training but neglect the inner group relations which can be used to divide the original group into multiple mini-groups. In order to overcome these obstacles, we design SAGD by dividing it into mini-groups according to the structure, which is more helpful to our encoder.

\section{Method}
\begin{figure*}
  \includegraphics[width=\textwidth]{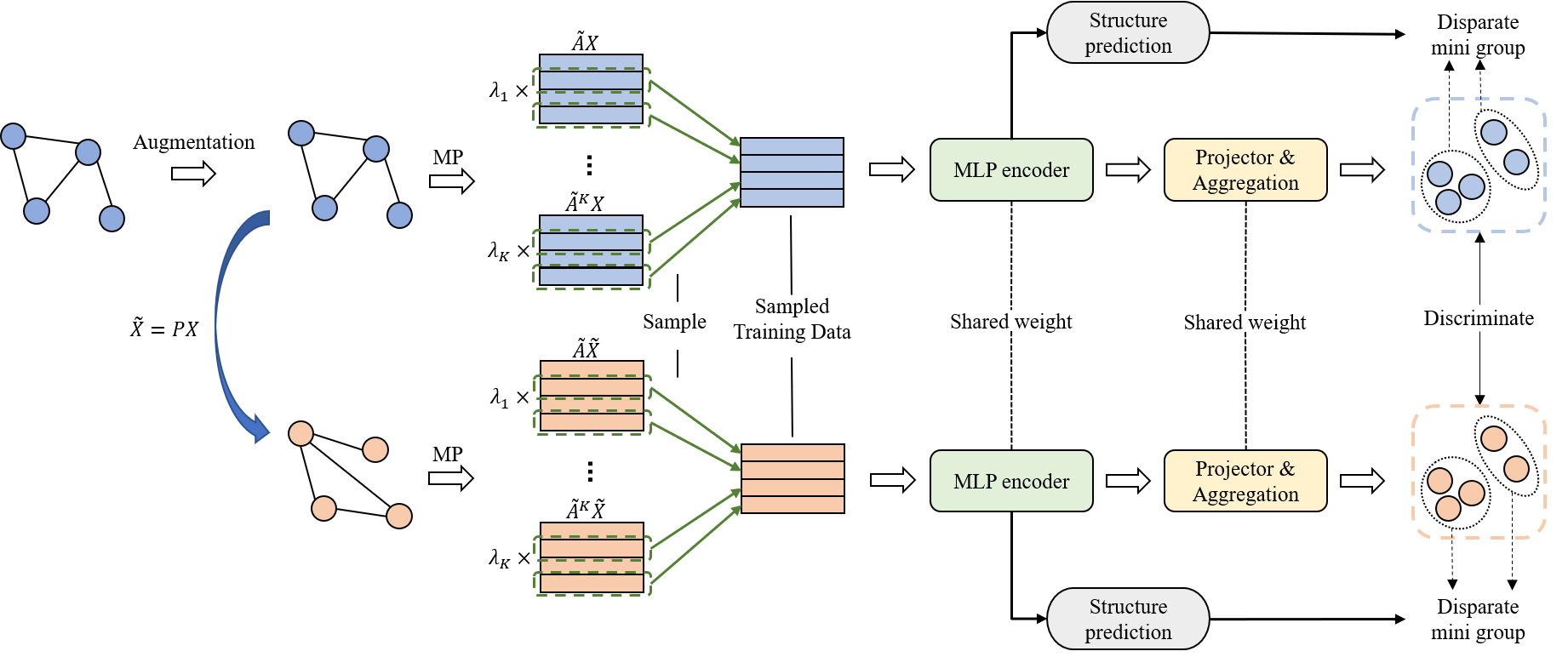}
  \caption{The architecture of AVGE-SAGD. Given a graph $\mathcal{G}$ and node attribute matrix $\mathbf{X}$, we first adopt an optional data augmentation and then generate negative samples by randomly permutating the node attributes matrix. Message passing is processed for both the positive sample and negative sample which will give $K$-views of features in $K$-hop. We sample $N/K$ features in each view to keep the scale of training data. The training data will be fed to the MLP encoder. After projection and aggregation, the generated embeddings can be discriminated into the positive group and negative group.}
  \label{fig:framework}
\end{figure*}

In this chapter, we will introduce our framework AVGE-SAGD in detail. The overall processes are shown in Figure~\ref{fig:framework}.

\subsection{Problem Formulation}

Given a graph $\mathcal{G}=(\mathbf{X}, \mathbf{A})$ with node attribute matrix $\mathbf{X}\in \mathbf{R}^{N\times d}$, where $N$ is the number of nodes, $d$ is node attribute dimension, and graph adjacency matrix $\mathbf{A}\in\mathbb{R}^{N\times N}$, where $A_{i,j}=1$ if node $i$ and $j$ are connected, else $A_{i,j}=0$. For message passing, we follow the setting in GCN, using normalized adjacency matrix $\tilde{\mathbf{A}}=\mathbf{D}^{-\frac{1}{2}}\mathbf{A}\mathbf{D}^{-\frac{1}{2}}$ where $\mathbf{D}$ is the diagonal degree matrix and $D_{ii}=d_i$ represent the number of degrees of node $i$. In our framework, we define the encoder as $f_\theta:\mathbb{R}^{N\times d}\rightarrow\mathbb{R}^{N\times d'}$ where $d'$ is the dimension of node representations.

The goal of GCL is to train a generalized graph encoder $f_\theta$ by a pretext loss $\mathcal{L}$ without labels. For evaluating the pre-trained model on a specific downstream task (\eg, node classification), we will obtain the node representations by the frozen encoder ($\mathbf{H}=f_\theta(\mathbf{A}, \mathbf{X})$). Then, we will train a linear classifier built on these node representations from the training set by a supervised loss (\eg, cross-entropy loss). Lastly, we will use the test set to evaluate the performance of our pre-trained model with the linear classifier.

\subsection{Generating Positive and Negative Samples}

\subsubsection{Data augmentation} 

We adopt data augmentation in generating positive samples. Node attribute masking is a popular technique and is widely used in GCL methods (\eg, GraphCL~\cite{you2020graph}, GGD~\cite{rethinking}). We adopt this data augmentation technique to enrich the features of positive samples. In practice, partial dimensions of node attributes will be masked with 0.

The augmented node attributes $\Tilde{\mathbf{X}}$ is obtained by:
\begin{equation}
    \Tilde{\mathbf{X}}=\mathbf{X} \circ \mathbf{M},
\end{equation}
where $\mathbf{M} \in \mathbb{R}^{N \times D}$ is masking matrix and each row vector in $\mathbf{M}$ are equal (\ie, $\mathbf{m}_i=\mathbf{m}_j,\forall{i,j}$), each element $m_{ij}$ in $\mathbf{m}_{i} \in \{0,1\}^{D}$ is is drawn from a Bernoulli distribution with probability $p_m$ (\ie, $m_{ij} \sim \mathcal{B}\left(1-p_m\right)$). In order to keep the notation uncluttered,

we use $\mathbf{X}$ to represent the augmented feature matrix in later sections.

\subsubsection{Corruption} We adopt corruption to generate negative samples. We randomly permutate the node attributes matrix and keep the topology structure unchanged:
\begin{equation}
    \check{g}=\{\check{\mathbf{X}},\mathbf{A}\}, \check{\mathbf{X}}=\mathbf{P}\mathbf{X},
\end{equation}
where $\mathbf{P}$ is a permutation matrix. 

This corruption technique 

is widely used in node-graph level GCL frameworks (\eg, DGI~\cite{DGI}, MVGRL~\cite{MVGRL}) to encourage the representations including structural similarities of different nodes in the graph properly. In our framework, this corruption operation will mislead message passing (\eg, $\mathbf{A}\check{\mathbf{X}}$) to generate erroneous node attributes as negative samples.

\subsection{Adaptive-View Graph Encoder}

\subsubsection{Post-message-passing features as training data}
With data augmentation and corruption, we can obtain multi-view features by parameters-free linear message passing:
$[\Tilde{\mathbf{A}}\mathbf{X};\Tilde{\mathbf{A}}^2\mathbf{X};...;\Tilde{\mathbf{A}}^K\mathbf{X}]$ and $[\Tilde{\mathbf{A}}\check{\mathbf{X}};\Tilde{\mathbf{A}}^2\check{\mathbf{X}};...;\Tilde{\mathbf{A}}^K\check{\mathbf{X}}]$ which will be used to train the encoder $f_\theta$. These features can be reused during training and inference which can save a lot of computational time.

Considering that we store $K$ views of attributes for each node by parameter-free linear message passing, the scale of training data of a graph with $N$ nodes increases from $N$ to $KN$ compared to standard GNN encoders, which is certainly contradictory to the goal of reducing computing time and memory. We use a simple sample method to make a trade-off between performance and computation cost. In each epoch, we randomly sample $\frac{N}{K}$ nodes to keep the input training data size as $N$, which is consistent with the standard GNN encoder training process.

Another alternative approach is to use the average or summation of features in different hops. Unfortunately, the information of different hops will be mixed up which leads to performance degeneration. However, our sample method can explicitly use features in more views that provide more distinct and useful information.

\subsubsection{Adaptive weighted training} \label{sec:adaptive}

Different hop features will be fed to train the encoder, but some hops' information is redundant and high-order hop features may incur oversmoothing~\cite{oversmoothing}. So, the contributions of each hop's feature should be disparate.

We assign individual adversarially learnable weight $\lambda_i$ to each hop feature $\mathbf{\Tilde{A}}^{i}\mathbf{X}$. The training data can be reformulated as

$[\lambda_1\mathbf{\Tilde{A}}\mathbf{X},\lambda_2\mathbf{\Tilde{A}}^2\mathbf{X},...,\lambda_K\mathbf{\Tilde{A}}^K\mathbf{X}]$.

In order to avoid the training loss easily converging to 0 (\ie, through minimizing training loss, discrepant high-order features will have large weights $\lambda_i$ and weights of indiscernible low-order will easily collapse to zero), we use a two-step min-max optimization method to train the MLP encoder with adaptive weighted multiple receptive field features. This training method can be formulated as:

\begin{align}
    \min_{\mathbf{\theta}} \max_{\mathbf{\lambda}} \quad & \mathcal{L}(\lambda_i\Tilde{\mathbf{A}}^i\mathbf{X},\lambda_i\Tilde{\mathbf{A}}^i\check{\mathbf{X}},\mathbf{\theta})\\
    \text{s.t.} \quad &
    \sum_i^K \lambda_i=1, \forall \lambda_i\in[0,1],
\end{align}
where $\lambda$ is Xavier initialised~\cite{glorot2010understanding} and $\mathcal{L}$ is our training loss which will be described in Section~\ref{sec:gd}. At each training step, firstly, we optimize adaptive weights with frozen model parameters by maximizing the training loss. Then, we optimize the parameters of the encoder with fixed adaptive weights by minimizing training loss. The optimization algorithm is described in Algorithm~\ref{alg:adaptive}.

\begin{algorithm}[tb]
    \caption{Adaptive weighted training algorithm}
    \label{alg:algorithm}
    \textbf{Input}: initial model parameter $\mathbf{\theta}^{(0)}$, adaptive weight $\mathbf{\lambda}^{(0)}$, total training epoch $E$\\
    \textbf{Parameter}: $\mathbf{\theta}$, $\mathbf{\lambda}$\\
    \textbf{Output}: Optimized model parameter $\mathbf{\theta}^{(N)}$ \\
    \begin{algorithmic}[1] 
        \FOR{$e=1$ to $E$}
        \STATE Maximization: fix $\mathbf{\theta}=\mathbf{\theta}^{(e-1)}$ and calculate the gradient of $\mathbf{\lambda}^{(e)}$
        \STATE Minimization: fix $\mathbf{\lambda}=\mathbf{\lambda}^{(e)}$ and calculate the gradient of $\mathbf{\theta}^{(e)}$
        
        \STATE update $\mathbf{\theta}$ and $\mathbf{\lambda}$
        \ENDFOR
        \STATE \textbf{return} Optimized parameter $\mathbf{\theta}^{(E)}$
    \end{algorithmic}
    \label{alg:adaptive}
\end{algorithm}

From a more theoretical aspect, our motivation is given from the analysis of research on homophilous graphs and heterophilous graphs~\cite{zhu2020beyond}. Homophily describes the similarity between adjacent nodes. The relevant studies~\cite{yan2021two} show that graph representation learning will benefit from message passing in a homophilous graph and the opposite in a heterophilous graph. The corruption operation disrupts the graph connection relationship, which will make the corrupted graph turn into a heterophilous graph. As the order of message passing hop increases, the node attributes in the positive group and negative group will be separated spontaneously. So the model without an adaptive weighted training method will take shortcuts by overly using high-order-hop features during pre-training, which will consequently cause the model to lose generalization ability. More details can be found in Appendix D.

\subsection{Structure-Aware Group Discrimination} \label{sec:gd}

\subsubsection{Group discrimination}

It has been empirically proven that the contrastive learning task in DGI can be transformed into a binary classification task named group discrimination~\cite{rethinking}. Following this method, we use a projector $g_\omega(\cdot)$ which consists of an MLP to map node representations into another latent space and then aggregate the projected representations. At last, we use binary cross entropy (BCE) loss to discriminate them into positive and negative groups, which are labeled as $y_i=1$ and $y_i=0$ respectively. The group discrimination loss can be formulated as:

\begin{align}
    \mathcal{L}_{\text{GD}}=-\frac{1}{2N} \sum_{i=1}^{2N}\left[y_i\log(\hat{y}_i)+(1-y_i)\log(1-\hat{y}_i)\right],
\end{align}
where $\hat{y}_i=agg(g_\omega(\mathbf{h}_i))$, $agg(\cdot)$ is summation aggregation.

\subsubsection{Preserving structure} 
\begin{figure}
  \includegraphics[width=\linewidth]{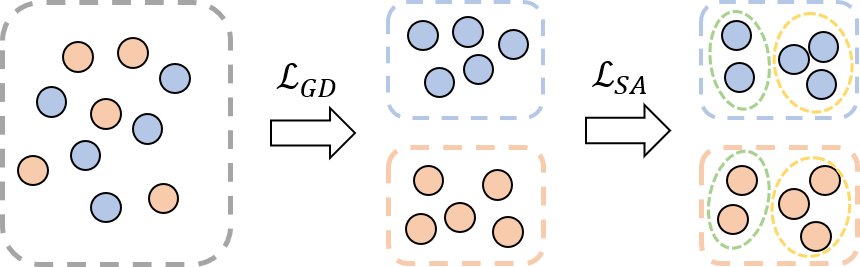}
  \caption{The schematic diagram of SAGD. $\mathcal{L}_\text{GD}$ means group discrimination loss and $\mathcal{L}_\text{SA}$ means structure aware loss. On the basis of $\mathcal{L}_\text{GD}$ distinguishing positive and negative samples, $\mathcal{L}_\text{SA}$ further distinguishes the mini-group according to the structure.}
  \label{fig:SAGD}
\end{figure}

Simple MLP encoder cannot preserve structure information~\cite{nosmog} because the training data are all independent node attributes in disparate hops. To solve this problem, we design auxiliary classification tasks to preserve structural information and capture the inner group relations so that the discriminated groups will be implicitly divided into mini-groups. Figure \ref{fig:SAGD} shows the procedure of SAGD.

Here we introduce the concept of relative degree which evaluates the node degree compared to its neighbors' degrees. The definition of the relative degree of node $v_i$ is:
\begin{equation}
    \bar{r}_i=\frac{1}{d_i}\sum_{j\in\mathcal{N}_i}\sqrt{\frac{d_i}{d_j}}.
\end{equation}

According to~\cite{yan2021two}, the nodes with a high relative degree are more sensitive to homophily and heterophily. In our case, the positive (high homophily) and negative (high homophily) samples generated by nodes with a high relative degree are highly discrepant. That is, the relative degree is a qualified graph structure indicator.

So encoders that can distinguish $\bar{r}$ preserve structural information and will therefore have stronger expressive power.

We hope that the formulation of the structure-preserving task can hold a consistent format of the discrimination task, which will be beneficial for model optimization.

Considering that relative degree is a continuous variable, we set $1$ as the threshold to discriminate whether a node has a high relative degree, which means $y_{\bar{r}_i}=1$ for a node with $\bar{r}_i > 1$ and $y_{\bar{r}_i}=0$ otherwise.

The relative degree loss can be written as:

\begin{align}
    \mathcal{L}_{\text{degree}}=-\frac{1}{2N}  \sum_{i=1}^{2N}\left[y_{\bar{r}_i}\log(\hat{y}_{\bar{r}_i})+(1-y_{\bar{r}_i})\log(1-\hat{y}_{\bar{r}_i})\right] ,
\end{align}
where $f_{\bar{r}}: \mathbb{R}^{D'}\rightarrow\mathbb{R}$ is a summation aggregation function and $\hat{y}_{\Bar{r}}$ is the prediction result.

Furthermore, we also use the hop order as the structural information that needs to be preserved. Similar to relative degree, we conduct a classification task to predict the order number of hop it belongs to through input features. The hop loss can be written as:
\begin{equation}
\begin{aligned}
    \mathcal{L}_{\text{hop}}=-\frac{1}{2N} &\sum_{i=1}^{2N}\left[y_{\text{hop}_i}\log(\hat{y}_{\text{hop}_i}) +  \right.\\ 
                                    & \left.(1-y_{\text{hop}_i})\log(1-\hat{y}_{\text{hop}_i}))\right],
\end{aligned}
\end{equation}
where $f_{\text{hop}}: \mathbb{R}^{D'}\rightarrow\mathbb{R}$ and $\hat{y}$ is the prediction result.

 \subsubsection{Final SAGD loss}
 Our structure-aware group discrimination loss can be written as:
\begin{equation}
    \mathcal{L}=\alpha\mathcal{L}_{\text{GD}}+\beta\mathcal{L}_{\text{hop}}+\gamma\mathcal{L}_{\text{degree}},
\end{equation}
where $\alpha,\beta,\gamma$ are hyper-parameters used for controlling the contributions of each loss. Empirically, we set $\alpha,\beta,\gamma$ as $1, 0.01, 0.05$ respectively in most cases.

\section{Experiments}

\begin{table*}[]
\renewcommand\arraystretch{1.15}
\centering
\setlength{\tabcolsep}{4mm}{
\begin{tabular}{c|c|ccccc}
\hline
\hline
                            &  Methods  & Cora     & CiteSeer & PubMed   & Computers     & Photo    \\
\hline
\hline
\multirow{2}{*}{Supervised}     & GCN     & $81.5$       & $70.3    $ & $79.0    $ & $76.3\pm0.5$ & $87.3\pm1.0$ \\
                                & GAT     & $83.0\pm0.7$ & $72.5\pm0.7$ & $79.0\pm0.3$ & $79.3\pm1.1$ & $86.2\pm1.5$ \\
                                 & SGC     & $81.0\pm0.0$ & $71.9\pm0.1$ & $78.9\pm0.0$ & $74.4\pm0.1$ & $86.4\pm0.0$ \\ \hline
\multirow{9}{*}{Self-supervised}& DGI & $81.7\pm0.6$ & $71.5\pm0.7$ & $77.3\pm0.6$ & $75.9\pm0.6$ & $83.1\pm0.5$ \\
                                & GMI     & $82.7\pm0.2$ & $73.0\pm0.3$ & $80.1\pm0.2$ & $76.8\pm0.1$ & $85.1\pm0.1$ \\
                                & MVGRL   & $82.9\pm0.7$ & $72.6\pm0.7$ & $79.4\pm0.3$ & $79.0\pm0.6$ & $87.3\pm0.3$ \\
                                & GRACE   & $80.0\pm0.4$ & $71.7\pm0.6$ & $79.5\pm1.1$ & $71.8\pm0.4$ & $81.8\pm1.0$ \\
                                & GraphCL & $82.5\pm0.2$ & $72.8\pm0.3$ & $77.5\pm0.2$ & OOM          & $79.5\pm0.4$ \\
                                & BGRL    & $80.5\pm1.0$ & $71.0\pm1.2$ & $79.5\pm0.6$ & $89.2\pm0.9$ & $91.2\pm0.8$ \\
                                & GBT    & $81.0\pm0.5$ & $72.8\pm0.2$ & $79.0\pm0.1$ & $88.5\pm1.0$ & $91.1\pm0.7$ \\
                                & GGD    & $83.9\pm0.4$ & $73.0\pm0.6$ & $81.3\pm0.8$ & $90.1\pm0.9$ & $92.5\pm0.6$ \\ \cline{2-7}
                                & Ours    &\pmb{$84.2\pm0.5$} & \pmb{$73.1\pm0.8$} & \pmb{$81.6\pm0.2$} & \pmb{$90.1\pm0.3$} & \pmb{$93.5\pm0.3$} \\ 
\hline
\hline
\end{tabular}}
\caption{Experiments results for node classification task on small-scale datasets. We report accuracy(\%) for all datasets. The best performance is in \textbf{bold}. OOM represents out-of-memory on NVIDIA GeForce RTX 3090 (24GB).}
\label{tab:normal datasets}
\end{table*}

In this section, we demonstrate that our framework AVGE-SAGD can achieve comparable performance in unsupervised representation learning for node classification with exceptional training and inference time. We evaluate the performance and computation time cost on various node classification datasets with the standard experiment settings.

\subsection{Datasets}

The datasets we use to evaluate our approach contain two types desperated by the data scale: small-scale datasets include Cora, CiteSeer, PubMed~\cite{sen2008collective}, Amazon Computers and Amazon Photo~\cite{shchur2018pitfalls} and large-scale datasets include ogbn-arxiv and ogbn-products provided by Open Graph Benchmark\cite{ogb}. Dataset statistics can be found in Appendix B.

In our implementation, we follow the standard data splits in~\cite{yang2016revisiting}. And for Amazon Computers and Photos, we randomly allocate $10/10/80\%$ of data to training/validation/test set respectively.

\subsection{Experimental Setup}

\subsubsection{Model}

For the encoder $f_\theta$, we use a 1-layer MLP for all datasets to save computing time. The projector $g_\omega$ is also a 1-layer MLP. $f_{\Bar{r}}$ and $f_\text{hop}$ are summation functions used for structure prediction.

\subsubsection{Inference}
During the inference phase, we freeze the trained MLP encoder $f_\theta$ and obtain final node representations $\mathbf{H}$ which can be used for downstream tasks with the processed input data. Since the input data comes from pre-processing, and the encoder is an MLP structure, the graph $\mathcal{G}$ is not needed in the inference stage, which saves a lot of computing resources in the message passing phase compared to the GNN encoder.

Different from the training step we use features in all hops as the training data, the final node representation is given from the features of the last hop only. In the training phase, we use the features of each hop. In order to maintain the consistency of the encoder input data, we only use one hop for inference in the inference stage. We choose the feature of the last hop because it contains the most neighbor information. Simple averaging of individual hop features will destroy high-order neighbor information. The final node representation is given by:
\begin{equation}
    \mathbf{H}=f_\theta(\Tilde{\mathbf{A}}^K\mathbf{X}),
\end{equation}
where $f_\theta$ is the MLP encoder, $\Tilde{A}$ is the normalized adjacency matrix of graph $\mathcal{G}$ and $\mathbf{X}$ is the original node attribute.

\subsubsection{Evaluation}

In our experiment, we evaluate the performance of our method by node classification tasks, following the most common GCL methods\cite{DGI,zhu2020deep,thakoor2021bootstrapped,zbontar2021barlow,peng2020graph,rethinking}. In detail, we train a simple logistic regression classifier by using the final node representations $\mathbf{H}$ and test the performance on the various node classification datasets. We measure the model performance using the averaged classification accuracy with ten results.

The evaluation of computation efficiency on large-scale datasets contains two parts: training efficiency and inference efficiency. Training efficiency is measured by the time spent per training epoch and inference efficiency is measured by the time spent for node embedding generation. We do not measure the time spent for classifying embeddings because we keep the complexity of the classifier the same. Note that in our framework, the message passing step does not need back propagation so that it can be separated from the encoder training procedure. This calculation can be done on other servers in a distributed system. Therefore we do not measure the time required for message passing in ogbn-arxiv. However, in ogbn-products, even if other servers are used to calculate message passing, the calculation time is still very long. So we count the time required to calculate message passing locally.

\subsubsection{Baselines}

First, we compare our framework with supervised GNNs (i.e., GCN~\cite{GCN}, GAT~\cite{GAT}, SGC~\cite{SGC}). Then we compare with some classical GCL methods (i.e., DGI~\cite{DGI}, MVGRL~\cite{MVGRL}, GRACE~\cite{GRACE}, GMI~\cite{GMI}, BGRL~\cite{BGRL}, GBT~\cite{GBT}). Finally, we compare a newly proposed efficient GCL method GGD~\cite{rethinking}. The reported results of some baselines are from previous papers if available.

\begin{figure*}[]
	\centering
	\begin{subfigure}{0.2\linewidth}
		\centering
		\includegraphics[width=0.7\linewidth]{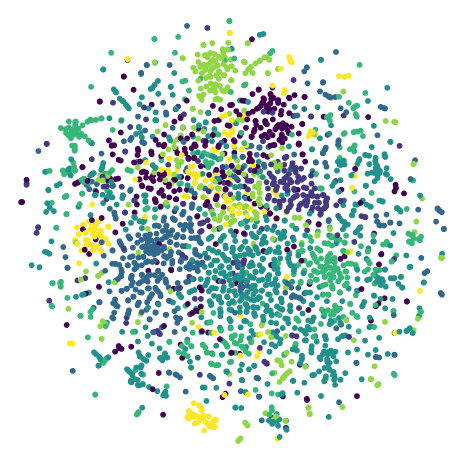}
		\caption{raw features}

	\end{subfigure}
	\centering
	\begin{subfigure}{0.2\linewidth}
		\centering
		\includegraphics[width=0.7\linewidth]{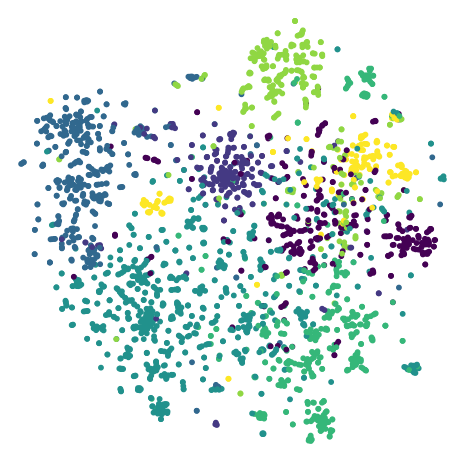}
		\caption{random-init}

	\end{subfigure}
    \centering
    \begin{subfigure}{0.2\linewidth}
		\centering
		\includegraphics[width=0.7\linewidth]{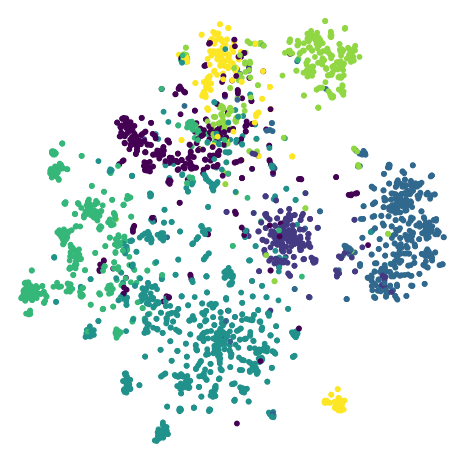}
		\caption{GGD}

	\end{subfigure}
	\centering
	\begin{subfigure}{0.2\linewidth}
		\centering
		\includegraphics[width=0.7\linewidth]{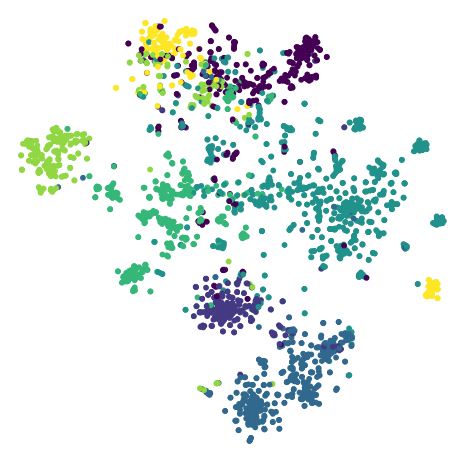}
		\caption{AVGE-SAGD}

	\end{subfigure}
	\caption{The t-SNE visualization result of node embeddings on Cora dataset. (a) is the raw features, (b) is the node embeddings from random initialized AVGE-SAGD, (c)is the learned representation of GGD, (d) is the learned representation of AVGE-SAGD.}
	\label{fig:tsne}
\end{figure*}

\subsection{Results and Analysis}
\subsubsection{Results for small-scale datasets} Table~\ref{tab:normal datasets} shows the classification results on five small-scale datasets, and we can draw some conclusions: (i) Experiment results show that our framework outperforms supervised GNNs and other state-of-the-art GCL baselines in all datasets, which shows the superiority of our AVGE-SAGD framework. (ii) Compared with GGD, our method surpasses it by a considerable margin (\eg, 1\% absolute improvement on Photo dataset) indicating the significance of structure-aware group discrimination. Our structure-aware group discrimination performs topology-based mini-group classification on the basis of graph group discrimination, which helps the model to learn more rich knowledge.

\subsubsection{Results for large-scale datasets} We evaluate the classification accuracy and computational efficiency of our model on two large-scale datasets provided by OGB\cite{ogb}: ogbn-arxiv and ogbn-products.

\begin{table}[]
\centering
\scalebox{0.85}{
\begin{tabular}{cccc}
\toprule
Methods  & Accuracy (\%) & \makecell[c]{Training\\Time (s)} & \makecell[c]{Inference\\Time (s)}\\
\midrule
GCN      & 71.7±0.3 &  -             &  -              \\
\midrule
MLP      & 55.5±0.2 &  -             &  -              \\
Node2vec & 70.1±0.1 &  -             &  -              \\
\midrule
DGI      & 70.3±0.2 &  /             &  /              \\
GRACE    & 71.5±0.1 &  /             &  /              \\
BGRL     & 71.6±0.1 &  /             &  /              \\
GBT      & 70.1±0.2 &  6.19             &  0.13              \\
GGD*      & 71.2±0.2 & 1.00          & 0.08           \\
\midrule
Ours     & 71.3±0.3 & 0.54          & 0.0003        \\
\bottomrule
\end{tabular}}
\caption{Accuracy on node classification task and speed test on the large-scale dataset ogbn-arxiv. 'Training Time' represents the average training time in each epoch. 'Inference Time' represents the time required from inputting data to computing the embedding. GGD* is the re-implementation on our devices with their official code. '/' means the method is OOM under a full-graph training setting.}
\label{tab:arxiv}
\end{table}

Experiment results in Table~\ref{tab:arxiv} and Table~\ref{tab:products} show that our framework has faster training speed and faster inference speed than most GCL frameworks, as well as GGD, which also uses group discrimination instead of pairwise contrastive learning paradigm. Although the result of our method is slightly lower than GRACE and BRGL in ogbn-arxiv, it saves a lot of computing resources and is memory-friendly. For ogbn-arxiv, we are 266 $\times$ faster than GGD in inference time and for ogbn-products we are 301 $\times$ faster. Since our message passing process does not contain parameters, our framework is still faster than the other GCL frameworks using GCN encoder. Due to the addition of auxiliary modules and tasks in our framework, which increases the number of additional calculations, the training speed improvement is relatively limited. But in the inference stage, 
the size of our model is equivalent to a simple MLP. So the inference efficiency has been greatly improved.

On the other hand, it is observed that on the large-scale dataset provided by OGB, the performance of GCL is inferior to the basic supervised GCN. The reason is there are plenty of training data on these datasets while the main contribution of GCL is the scenario lacking training data, so it cannot performs better than supervised models on these datasets. In the small-scale datasets with very limited training data mentioned in the last paragraph, however, the overall performance of GCL is significantly improved compared with the supervised models.

\begin{table}[]
\centering
\scalebox{0.85}{
\begin{tabular}{cccc}
\toprule
Methods  & Accuracy (\%) & \makecell[c]{Training\\Time (s)} & \makecell[c]{Inference\\Time (s)}\\
\midrule
GCN      & 75.6±0.2 & -             & -              \\
\midrule
MLP      & 61.1±0.0 & -             & -              \\
Node2vec & 68.8±0.0 & -             & -              \\
\midrule
BGRL     & 64.0±1.6 & 2267             & 265              \\
GBT      & 70.5±0.4 & 1963             & 262              \\
GGD*(1024)     & 75.6±0.2 & 779      &  718              \\
GGD*(256)     & 73.3±0.4 & 555      &  301              \\
\midrule
Ours(256)     & 75.9±0.1 & 5 (364)           &  1          \\
\bottomrule
\end{tabular}}
\caption{Accuracy on node classification task and speed test on the large-scale dataset ogbn-products. In the method column, the number in the brackets means the dimension of embeddings. In the accuracy column, the number in the brackets means the training time with message passing. GGD* is the re-implementation on our devices with their official code.}
\label{tab:products}
\end{table}

\subsection{Ablation Study} \label{sec:ab}

To prove the effectiveness of the design module of our framework, we conduct ablation experiments with different modules under the same hyperparameters on Cora dataset. More results are in Appendix C. In Table~\ref{tab:ablation}, `Multi-View Weights' includes different strategies for adopting weights on multi-view attributes by masking different components. The first three rows assign fixed weights to different hop attributes. $[0,0,...,1]$ means we only use the attributes of the last hop to train the encoder. $[1,1,...,1]$ means we keep the contributions of different hop attributes the same. The last three columns represent that we use learnable weights to adjust weights adaptively. `min' represents that we optimize weights and model parameters by minimizing training loss. `min-max' represents that we use a two-step adaptive weighted training method, `Structure Preserving' means structure-aware module.

The results show that all of the modules we design are helpful for the performance of our framework. The two-step min-max adaptive weight training method is the most significant part in the framework since the performance degrades without it. And with structure-preserving module, SAGD outperforms GGD in our framework. Furthermore, we observe that using fixed multi-hop feature training performs worse than using the last-hop feature only, which underscores the importance of our adaptive weighted training approach.

\begin{table}[]
\renewcommand\arraystretch{1.15}
\begin{tabular}{c|ccc}
\Xhline{0.8pt}
                                 & Multi-View Weights& \makecell[c]{Structure \\Preserving} & \makecell[c]{Accuracy\\(on Cora)} \\ \hline
\multirow{3}{*}{\makecell[c]{Fixed \\Weights}} & $\lambda=[0,0,...,1]$   &                      & 83.5±0.3          \\ \cline{2-4}  
                                 & $\lambda=[1,1,...,1]$   &                      & 83.4±0.4          \\ \cline{2-4} 
                                 & $\lambda=[1,1,...,1]$   & \CheckmarkBold       & 83.5±0.7          \\ \hline 
\multirow{3}{*}{\makecell[c]{Learnable \\Weights}}& $\min_\theta\mathcal{L}$             &                      & 83.6±0.4          \\ \cline{2-4} 
                                 & $\min_\theta$-$\max_\lambda\mathcal{L}$         &                      & 84.0±0.6          \\ \cline{2-4} 
                                 & $\min_\theta$-$\max_\lambda\mathcal{L}$          & \CheckmarkBold       & 84.2±0.5          \\ \Xhline{0.8pt} 
\end{tabular}
\caption{Ablation studies for AVGE-SAGD on Cora dataset.}
\label{tab:ablation}
\end{table}

\subsection{Visualization}

To visually assess the quality of our learned embeddings, we adopt t-SNE~\cite{tsne} to visualize the 2D projection of node embeddings on Cora dataset using raw features, random-init of AVGE-SAGD, GGD, and trained AVGE-SAGD in Figure~\ref{fig:tsne}, where nodes in different labels have different colors.

We can observe that the distribution of node embeddings in raw features and random-init are messy and intertwined. After training, node embeddings learned by AVGE-SAGD have a clear separation of clusters, which indicates the model can help learn representative features for downstream tasks. Compared to GGD, the margins of each cluster of node embeddings learned from AVGE-SAGD are much wider, which means higher quality.

\section{Conclusion}

In this paper, we approach to the challenge of increasing the training and inference efficiency of the graph contrastive representation learning frameworks. In terms of improving the encoder's efficiency, we separate the message passing from the embedding prediction and design a novel adversarially adaptive weights multi-hop features. As for the pre-training loss, we built a new structure-aware group discrimination loss that helps our fast encoder to preserve more structural information, which consequently improves its generalization ability on the downstream tasks. Extensive experiments conducted on both small-scale and large-scale datasets have shown the effectiveness of our framework regarding both downstream task performance and the training and inference speed.

\bibliographystyle{named}
\bibliography{ref}

\newpage

\appendix
\section{Algorithm}
The steps of the procedure of our method are summarised as:
\begin{algorithm}[ht]
    \caption{Algorithm for structure-aware group discrimination with adaptive-view graph encoder}
    \label{alg:framework}
    \textbf{Input}: Graph $\mathcal{G}=(\mathbf{X},\mathbf{A})$, encoder $f_{\theta}$, projector $g_{\omega}$, structure preserver $f_{\bar{r}}$ and $f_{\text{hop}}$, training step $E$. \\

    \begin{algorithmic}[1] 
        \STATE Data augmentation and corruption. Augmented positive graph samples: $\{\mathcal{G}_i = (\mathbf{A}, \mathbf{X})\}$ and corrupted negative graph samples: $\{\check{\mathcal{G}}_i = (\mathbf{A}, \check{\mathbf{X}})\}$, where $i\in \mathbb{N}^E$
        \STATE Linear message passing for positive graph samples $\{\mathbf{D}_i=[\Tilde{\mathbf{A}}\mathbf{X};\Tilde{\mathbf{A}}^2\mathbf{X};...;\Tilde{\mathbf{A}}^K\mathbf{X}]_i\}$ and negative graph samples: $\{\check{\mathbf{D}}_i=[\Tilde{\mathbf{A}}\check{\mathbf{X}};\Tilde{\mathbf{A}}^2\check{\mathbf{X}};...;\Tilde{\mathbf{A}}^K\check{\mathbf{X}}]_i\}$, where $i\in \mathbb{N}^E$
        \FOR{$e=1$ to $E$}
            \STATE Sample $\frac{N}{K}$ node to obtain $\mathbf{D}'$ and $\check{\mathbf{D}}'$ from $\mathbf{D}$ and $\check{\mathbf{D}}$
            \STATE Concatenate $\mathbf{D}'$ and $\check{\mathbf{D}}'$ to obtain $\bar{\mathbf{D}}$
            \STATE Assign adaptive weight $\lambda$ to $\bar{D}$
            \STATE Input $\bar{\mathbf{D}}$ to obtain node embeddings $\mathbf{H}=f_\theta(\bar{\mathbf{D}})$
            \STATE obtain the group discrimination prediction vector $\hat{\mathbf{y}}=agg(g_\omega(\mathbf{H})) \in \mathbb{R}^{2N}$
            \STATE obtain the structure-aware prediction vector $\hat{\mathbf{y}}_{\bar{r}}=f_{\bar{r}}(\mathbf{H}))\in \mathbb{R}^{2N}$ and $\hat{\mathbf{y}}_{\text{hop}}=f_{\text{hop}}(\mathbf{H}))\in \mathbb{R}^{2N}$
            \STATE Compute group discrimination loss $\mathcal{L}_{\text{GD}}$
            \STATE Compute structure-aware loss $\mathcal{L}_{\text{degree}}$ and $\mathcal{L}_{\text{hop}}$
            \STATE Compute the final loss $\mathcal{L}=\alpha\mathcal{L}_{\text{GD}}+\beta\mathcal{L}_{\text{hop}}+\gamma\mathcal{L}_{\text{degree}}$
            \STATE Update trainable parameters using adaptive weighted training algorithm in Algorithm 1

        \ENDFOR
        \STATE Obtain final embeddings $\mathbf{H}=f_\theta(\mathbf{A}^K\mathbf{X})$
    \end{algorithmic}
\end{algorithm}

\section{Experiment Details}

\subsubsection{Dataset statistics}

We present the details of node classification datasets we used in Table~\ref{tab:datasets}.

\begin{table}[ht]
\centering
\scalebox{0.85}{
\begin{tabular}{ccccc}
\toprule
Datasets         & Nodes     & Edges      & Features & Classes \\
\midrule
Cora             & 2,708     & 5,429      & 1,433    & 7       \\
Citeseer         & 3,327     & 4,732      & 3,703    & 6       \\
PubMed           & 19,717    & 44,338     & 500      & 3       \\
Amazon Computers & 13,752    & 245,861    & 767      & 10      \\
Amazon Photo     & 7,650     & 119,081    & 745      & 8       \\
ogbn-arxiv       & 169,343   & 1,166,243  & 128      & 40      \\
ogbn-products    & 2,449,029 & 61,859,140 & 100      & 47     \\
\bottomrule

\end{tabular}}

\caption{Datasets statistics}
\label{tab:datasets}
\end{table}

\subsubsection{Computer infrastructures specifications}

For hardware, we conduct all experiments on a computer server with eight GeForce RTX 3090 GPUs with 24GB memory and 64 AMD EPYC 7302 CPUs. Besides, our code is implemented based on PyTorch 1.12.1 and DGL0.9.1.

\subsubsection{Hyperparameters}

Among all hyperparameters, we consider hidden size, hop order, learning rate, $\alpha$, $\beta$, and $\gamma$ which are listed in Table~\ref{tab:hyper}, where $\alpha$, $\beta$ and $\gamma$ are the weight of loss discussed in Section 3.4

\begin{table}[ht]
\scalebox{0.85}{
\begin{tabular}{ccccccc}
\toprule
Dataset           & \begin{tabular}[c]{@{}c@{}}Hidden\\ Size\end{tabular} & \begin{tabular}[c]{@{}c@{}}Hop\\ Order\end{tabular} & \begin{tabular}[c]{@{}c@{}}Learning\\ Rate\end{tabular} & $\alpha$ & $\beta$    & $\gamma$    \\
\midrule
Cora             & 512                                                   & 2                                                   & 1e-3                                                    & 1 & 0.01 & 0.05 \\
Citeseer         & 1024                                                  & 1                                                   & 5e-4                                                    & 1 & 0.01 & 0.05 \\
PubMed           & 1024                                                  & 2                                                   & 1e-3                                                    & 1 & 0.01 & 0.05 \\
\makecell[c]{Amazon\\ Computers} & 1024                                                  & 2                                                   & 5e-4                                                    & 1 & 0.01 & 0.05 \\
\makecell[c]{Amazon\\ Photos}    & 512                                                   & 2                                                   & 1e-4                                                    & 1 & 0.01 & 0.02 \\
ogbn-arxiv       & 1500                                                  & 3                                                   & 5e-5                                                    & 1 & 0.01 & 0.05 \\
ogbn-products    & 256                                                   & 5                                                   & 5e-5                                                    & 1 & 0.01 & 0.05\\
\bottomrule
\end{tabular}}
\caption{Parameter settings on seven datasets}
\label{tab:hyper}

\end{table}

\section{Additional Experiments}

We conduct more ablation experiments on five small and medium datasets following the setting in Section 4.4. The results in Table~\ref{tab:addablation} show that our AVGE module and SAGD module can improve the performance separately, and the model achieves the best performance when we use both AVGE and SAGD techniques together.

\begin{table*}[ht]
\centering
\renewcommand\arraystretch{1.15}
\begin{tabular}{c|ccccccc}
\Xhline{0.8pt}
                                 & Multi-View Weights& \makecell[c]{Structure \\Preserving} & Cora & Citeseer & PubMed & Comp &Photo\\
                                 \hline
\multirow{3}{*}{\makecell[c]{Fixed \\Weights}} & $\lambda=[0,0,...,1]$   &                   & 83.5±0.3 & 71.7±0.7&  80.9±0.5 & 89.9± 0.2 & 92.8±0.3      \\ 
\cline{2-8}  
                                 & $\lambda=[1,1,...,1]$   &                      & 83.4±0.4   &71.7±0.4 & 81.0±0.4  & 89.6±0.4 & 92.8±0.4   \\ 
\cline{2-8} 
                                 & $\lambda=[1,1,...,1]$   & \CheckmarkBold       & 83.5±0.7   &71.8±0.4 & 81.2±0.5  & 90.0±0.2 & 93.2±0.1   \\ 
                                 \hline 
\multirow{3}{*}{\makecell[c]{Learnable \\Weights}}& $\min_\theta\mathcal{L}$             &                      & 83.6±0.4   &71.8±0.4 & 81.2±0.4 & 90.0±0.3 & 93.2±0.2    \\ 
\cline{2-8} 
                                 & $\min_\theta$-$\max_\lambda\mathcal{L}$         &                      & 84.0±0.6    &71.9±0.5 & 81.3±0.3 & 90.1±0.2 & 93.3±0.3   \\ 
\cline{2-8} 
                                 & $\min_\theta$-$\max_\lambda\mathcal{L}$          & \CheckmarkBold       & \textbf{84.2±0.5}   &\textbf{72.0±0.3} & \textbf{81.4±0.1} & \textbf{90.2±0.2} &  \textbf{93.5±0.2}   \\ 
                                 \Xhline{0.8pt} 
\end{tabular}
\caption{Ablation studies for AVGE-SAGD}
\label{tab:addablation}
\end{table*}

\section{More on Homo/Hetero-phily and Multi-Hop Features}

\begin{figure*}[ht]
	\centering

	\begin{subfigure}{0.2\linewidth}
		\centering
		\includegraphics[width=0.7\linewidth]{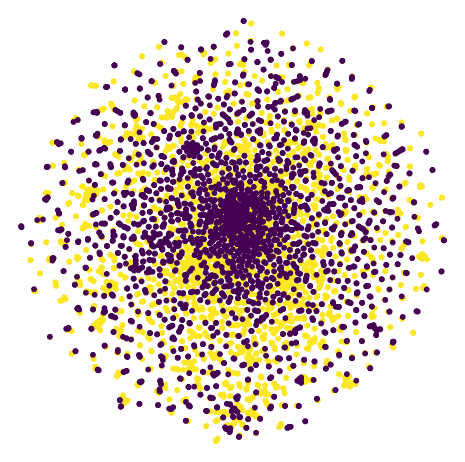}
		\caption{cora 0 hop}

	\end{subfigure}
 \begin{subfigure}{0.2\linewidth}
		\centering
		\includegraphics[width=0.7\linewidth]{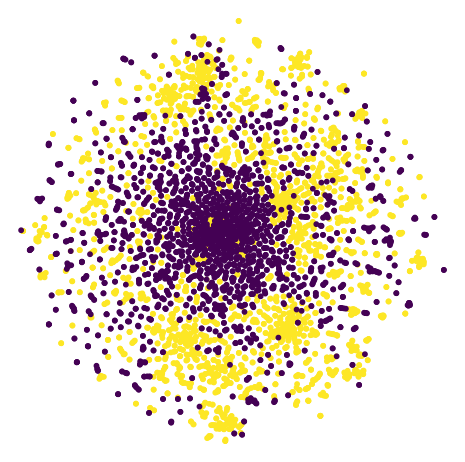}
		\caption{cora 1 hop}
		\label{cora 1 hop}
	\end{subfigure}
 \begin{subfigure}{0.2\linewidth}
		\centering
		\includegraphics[width=0.7\linewidth]{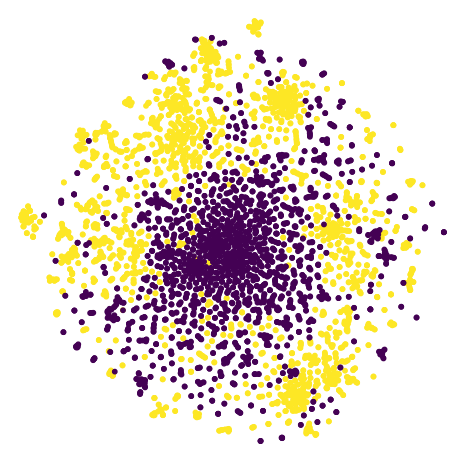}
		\caption{cora 2 hop}
		\label{cora 2 hop}
	\end{subfigure}
 \begin{subfigure}{0.2\linewidth}
		\centering
		\includegraphics[width=0.7\linewidth]{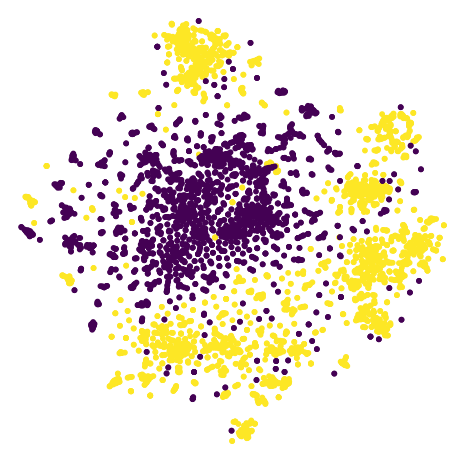}
		\caption{cora 8 hop}
		\label{cora 8 hop}
	\end{subfigure}

	
	\	\begin{subfigure}{0.2\linewidth}
		\centering
		\includegraphics[width=0.7\linewidth]{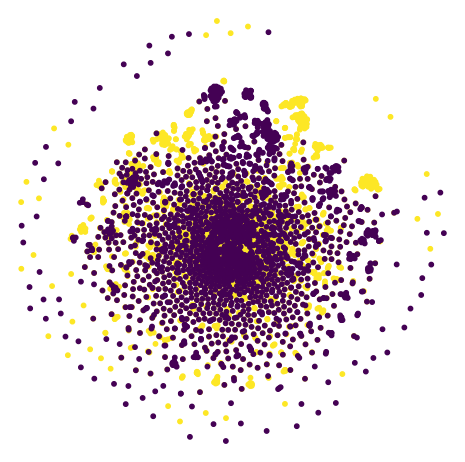}
		\caption{chameleon 0 hop}
		\label{chameleon 0 hop}
	\end{subfigure}
 \begin{subfigure}{0.2\linewidth}
		\centering
		\includegraphics[width=0.7\linewidth]{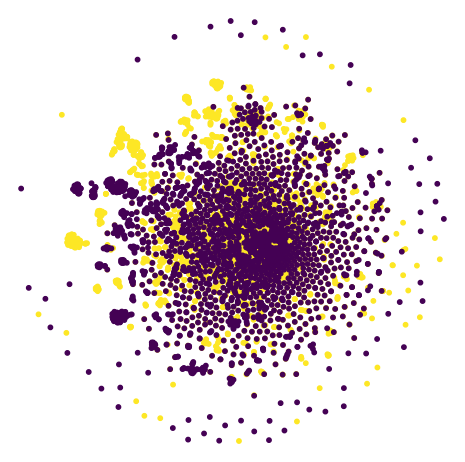}
		\caption{chameleon 1 hop}
		\label{chameleon 1 hop}
	\end{subfigure}
 \begin{subfigure}{0.2\linewidth}
		\centering
		\includegraphics[width=0.7\linewidth]{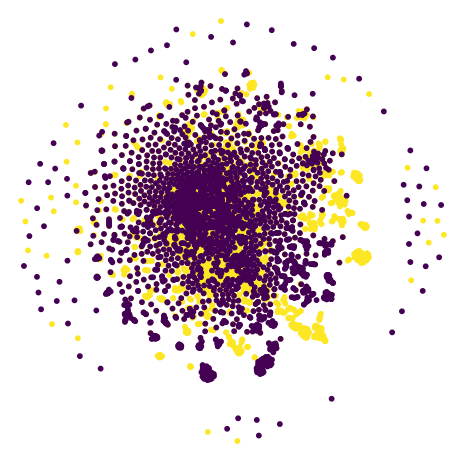}
		\caption{chameleon 2 hop}
		\label{chameleon 2 hop}
	\end{subfigure}
 \begin{subfigure}{0.2\linewidth}
		\centering
		\includegraphics[width=0.7\linewidth]{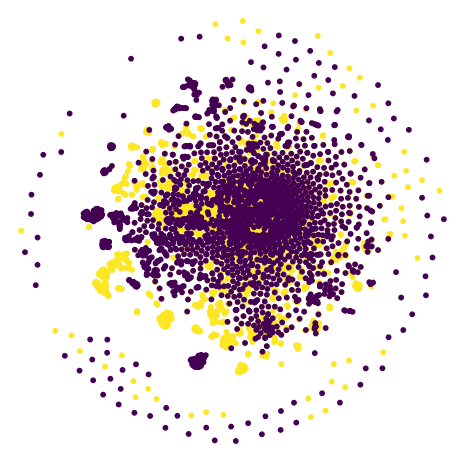}
		\caption{chameleon 8 hop}
		\label{chameleon 8 hop}
	\end{subfigure}
    \caption{Results of message passing in Cora (above) and Chameleon (below). The purple nodes are positive samples and the yellow nodes are negative samples.}
    \label{fig:MPret}
\end{figure*}

In this section, we will discuss the relationship between homophily and adaptive-view graph encoder and group discrimination.

Homophily of a graph is typically defined based on the probability of edges connection between nodes within the same class in supervised node classification datasets~\cite{zhu2020beyond}. The homophily ratio of edges is defined as 
\begin{align}
    h=\frac{1}{E}\sum_{(i,j)_\in E} \mathbf{1}(y_i=y_j) ,
\end{align}
where $|E|$ is the number of edges, $\mathbf{1}(\cdot)$ denotes the indicator function. Homophily characterizes the properties of the connection of nodes within the same class in graphs. Nodes in graphs with high homophily tend to be connected to nodes of the same class, while nodes in graphs with low homophily tend to be connected to nodes of different classes which is also called heterophily. 

Homophily has an important impact on the performance of message-passing-based GNNs. On graphs with high homophily, message passing can improve the expressive ability of node representations while on graphs with low homophily, message passing will reduce the expressive ability of node representations instead. On some heterophilous graph datasets, the performance of GCN is not as good as MLP which directly uses node features~\cite{zhu2020beyond}.

The method we use to construct negative samples through permutation can be seen as disrupting the connection relationship of the graph, which means that the homophily of the graph becomes very low. Figure \ref{fig:MPret} shows the disparate results of message passing in homophilous and heterophilous graphs with the number of hops from $0$ to $8$. On the homophilous dataset (Cora, $h=0.81$), with the increase of message passing order, the node features are divided into obvious positive and negative sample clusters from the initially mixed state, while on the hetreophilous dataset (Chameleon $h=0.32$)~\cite{pei2020geom}, the node features are mixed together from start to finish. 

In our framework, the random permutation corruption method changes the node connection relationship while preserving the graph's global topology. So a graph with high homophily will degrade to a graph with low homophily after corruption. Therefore, on the homophilous graph, the difference between positive and negative samples will increase with the increase of the number of message-passing layers and finally lead to the spontaneous formation of two clusters. Features in higher-order hops will perform better in group discrimination pretext task but are not conducive to model learning because the loss is too low. This observation indicates the necessity of our two-step min-max optimization training method in Section 3.3, aiming to automatically control the significance of features in the different hops.

\end{document}